\pdfoutput=1

\documentclass[11pt]{article}
\usepackage{textgreek}
\usepackage{marginnote}

\usepackage{color,soul}

\DeclareRobustCommand{\hlsnow}[1]{{\sethlcolor{snowgray!15!white}\hl{#1}}}

\newcommand{\sidenote}[2][10px]{%
  \marginnote{%
    \begingroup
    \color{snowgray}\sffamily\bfseries
    \rule{0pt}{#1}
    \ignorespaces#2\unskip
    \endgroup
  }%
}

\DeclareRobustCommand{\hlsnowtwo}[1]{{\sethlcolor{snowpurple!15!white}\hl{#1}}}

\newcommand{\sidenotetwo}[2][10px]{%
  \marginnote{%
    \begingroup
    \color{snowpurple}\sffamily\bfseries
    \rule{0pt}{#1}
    \ignorespaces#2\unskip
    \endgroup
  }%
}

\DeclareRobustCommand{\hlsnowthree}[1]{{\sethlcolor{snowpink!15!white}\hl{#1}}}

\newcommand{\sidenotethree}[2][10px]{%
  \marginnote{%
    \begingroup
    \color{snowpink}\sffamily\bfseries
    \rule{0pt}{#1}
    \ignorespaces#2\unskip
    \endgroup
  }%
}

\makeatletter
\AtBeginDocument{%
  \begingroup
  \normalsize
  \let\tmp@n@s\f@size
  \let\tmp@n@b\f@baselineskip
  \small
  \let\tmp@s@s\f@size
  \let\tmp@s@b\f@baselineskip
  \xdef\semismall@size{\fpeval{(\tmp@n@s+\tmp@s@s)/2}}%
  \xdef\semismall@baselineskip{\fpeval{(\tmp@n@b+\tmp@s@b)/2}}%
  \endgroup
}
\newcommand{\semismall}{\fontsize{\semismall@size}{\semismall@baselineskip}\selectfont}

\usepackage{booktabs}

\usepackage{textalpha}
\usepackage{tcolorbox}
\usepackage{todonotes}
\usepackage{tabularx}
\usepackage{amssymb}
\usepackage{wrapfig}
\usepackage{enumitem}
\usepackage[final]{acl}

\usepackage{times}
\usepackage{latexsym}
\usepackage{bibentry}
\usepackage{csquotes}
\usepackage[T1]{fontenc}

\usepackage[utf8]{inputenc}

\usepackage{microtype}
\usepackage{float}

\usepackage{inconsolata}

\usepackage{graphicx}
\usepackage{tikz}
\usetikzlibrary{patterns}
\definecolor{snowpurple}{HTML}{7254A3}
\definecolor{snowlight}{HTML}{7AC5EA}
\definecolor{snowdark}{HTML}{1F4463}
\definecolor{snoworange}{HTML}{FF9F36}
\definecolor{snowpink}{HTML}{D45B90}
\definecolor{snowgray}{HTML}{5B5B5B}
\definecolor{snowstar}{HTML}{75CDD7}

\definecolor{gray}{HTML}{BDBDBD}

\newtcolorbox{postulate}[1][]{
    colframe=yellow,
    colback=yellow!30!white,
    sharp corners,
    boxsep=0pt,
    left=5pt,
    right=5pt,
    top=6pt,
    bottom=6pt,
    boxrule=0pt,
    leftrule=4pt,
    #1
}

\newtcolorbox{translationtwo}[1][]{
    colframe=snowpurple!13!white,
    colback=snowpurple!13!white,
    sharp corners,
    boxsep=5pt,
    left=5pt,
    right=5pt,
    top=6pt,
    bottom=6pt,
    boxrule=0pt,
    leftrule=0pt,
    fontupper=\semismall,
    #1
}

\newtcolorbox{translation}[1][]{
    colframe=gray,
    colback=gray!15!white,
    sharp corners,
    boxsep=5pt,
    left=5pt,
    right=5pt,
    top=6pt,
    bottom=6pt,
    boxrule=0pt,
    leftrule=0pt,
    fontupper=\semismall,
    #1
}

\usepackage{newfloat}

\DeclareFloatingEnvironment[fileext=loa,name=Example,placement=!b]{examplefloat}

\newtcolorbox{examplebox}[1][]{
  colframe=snowgray!90!white,
  colback=snowgray!5!white,
  sharp corners,
  boxsep=0pt,
  left=5pt,
  right=5pt,
  top=5pt,
  bottom=6pt,
  boxrule=2pt,
  fonttitle=\bfseries,
  title=Example,
  coltitle=white,
  fontupper=\semismall,
  toptitle=3pt,      
  bottomtitle=3pt,   
  #1
}

\newtcolorbox{exampleboxtwo}[1][]{
  colframe=snowpurple!90!white,
  colback=snowpurple!5!white,
  sharp corners,
  boxsep=0pt,
  left=5pt,
  right=5pt,
  top=5pt,
  bottom=6pt,
  boxrule=2pt,
  fonttitle=\bfseries,
  title=Example,
  coltitle=white,
  fontupper=\semismall,
  toptitle=3pt,      
  bottomtitle=3pt,   
  #1
}

\definecolor{worldknowledge}{HTML}{E9E4F6}
\definecolor{practical}{HTML}{FDF1E4}
\definecolor{complex}{HTML}{E6F4Fb}
\definecolor{resistance}{HTML}{E1F2DB}

\newlength{\shlDesiredTopBottomPad}
\newlength{\shlDesiredSidePad}
\newlength{\shlOriginalFboxsepForReporting}
\DeclareRobustCommand{\shl}[3]{%
  \mbox{
    \begingroup%
      \def\shlcoretext{{\vphantom{Ay}#2}}%
      \sbox0{\shlcoretext}%
      \dimen0=\ht0%
      \advance\dimen0 by \shlOriginalFboxsepForReporting%
      \dimen1=\dp0%
      \advance\dimen1 by \shlOriginalFboxsepForReporting%
      \dimen2=\shlDesiredSidePad%
      \advance\dimen2 by -\shlDesiredTopBottomPad%
      \def\shlfullboxcontent{{\hspace*{\dimen2}\shlcoretext\hspace*{\dimen2}}}%
      \setlength{\fboxsep}{\shlDesiredTopBottomPad}%
      \raisebox{0pt}[\dimen0][\dimen1]{%
        \colorbox{#1}{\shlfullboxcontent}%
      }%
    \endgroup%
  }
  \ 
}

\definecolor{darkworldknowledge}{HTML}{7547C8}
\definecolor{darkpractical}{HTML}{FF9F36}
\definecolor{darkcomplex}{HTML}{E6F4Fb}
\definecolor{darkresistance}{HTML}{E1F2DB}

\title{Language Models Model Language}

\author{Łukasz Borchmann \\ Snowflake AI Research \\ \texttt{lukasz.borchmann@snowflake.com}}

\begin{document}
\maketitle
\begin{abstract}
Linguistic commentary on LLMs, heavily influenced by the theoretical frameworks of de Saussure and Chomsky, is often speculative and unproductive.
Critics challenge whether LLMs can legitimately model language, citing the need for ``deep structure'' or ``grounding'' to achieve an idealized linguistic ``competence.'' We argue for a radical shift in perspective towards the empiricist principles of Witold Mańczak, a prominent general and historical linguist. He defines language not as a ``system of signs'' or a ``computational system of the brain'' but as the totality of all that is said and written. Above all, he identifies frequency of use of particular language elements as language's primary governing principle. Using his framework, we challenge prior critiques of LLMs and provide a constructive guide for designing, evaluating, and interpreting language models.
\end{abstract}

\section{Introduction}

For two thousand years, language scholars have produced myriad works without clearly defining a way to validate their claims \cite{Manczak1981KryteriaOrg,Manczak1996Kryteria}. This lack of rigor left the field uniquely unprepared for the empirical success of Large Language Models (LLMs).

\looseness=-1 When GPT or Claude synthesizes complex responses, most users of LLMs perceive this ability as a sign of linguistic competence. At the same time, orthodox linguists 
rush to claim that apparent ``competence'' is merely an illusion, because LLMs' architecture and training techniques do not fit established theories of language.
These objections reveal more about current linguistic dogma than about the capabilities of LLMs.

Debate remains anchored in theoretical frameworks inherited from de Saussure's abstract ``system of signs,'' Chomsky's postulated innate mental grammar, and even Plato's conception of language as a communication tool. When scholars criticize LLMs for failing to ``explain the rules of English syntax'' \cite{chomsky_false_2023}, point out that they don't distinguish ``between correctness and likelihood'' \cite{FoxKatzir}, or question whether they can ``have access to meaning'' despite experiencing only form \cite{10.1145/3442188.3445922}, these researchers apply criteria derived from specific---and contestable---theories. Such critiques function less as objective assessments of LLMs' capacities than defenses of a linguistic paradigm that has been challenged by new empirical evidence.

\looseness=-1 We propose a radical theoretical reorientation and turn to Witold Mańczak (1924-2016), whose critique of abstract, dualistic linguistics offers an alternative foundation. 
While \citet{10.1145/3442188.3445922} advocate constraining the LLMs' tendency to act like ``stochastic parrots,'' we call for a new science of ornithology that is equipped to understand what has actually taken flight.

\section{The Mańczakian Framework}

The Mańczakian framework is best understood as a response to what its author saw as a foundational crisis in linguistics: the absence of scientific criteria for determining ``truth.'' Mańczak argued that the field had long substituted the authority of famous scholars for the authority of verifiable evidence \cite{Manczak1980,Manczak1981KryteriaOrg,Manczak1982,Manczak1988,Manczak1996Kryteria}.

\sidenote{Science of Language}The central goal of his work was to \hlsnow{reground linguistics as an inductive, quantitative science}. He insisted that any valid claim must be subject to statistical or experimental verification and explained by observable phenomena rather than abstract speculation. Applying quantitative methods to vast corpora, he discovered that frequency of use of language elements is a primary force shaping language.

We will now unpack these core tenets, demonstrating how the empiricist lens of Mańczak reframes long-standing linguistic debates and provides a robust foundation for the science of language and language models.

\begin{table*}[t!]
\centering
\semismall
\caption{Comparison of dominant and Mańczakian linguistic paradigms.}
\label{tab:paradigm_shift}
\begin{tabularx}{\textwidth}{>{\raggedright\arraybackslash}p{3.4cm} >{\raggedright\arraybackslash}X >{\raggedright\arraybackslash}X}
\toprule
\textbf{Aspect} & \textbf{Dominant Paradigms} & \textbf{Mańczakian Paradigm} \\
\midrule
\textbf{Definition of~Language} & An abstract ``system of signs'' (de Saussure) or mental ``competence'' distinct from performance (Chomsky). & The material totality of all that is said and written---the actual corpus of texts. \\
\addlinespace
\textbf{Organizing Principle} & Innate universal grammar; binary structural oppositions; algebraic rules for generating well-formed sentences. & Frequency of use. High-frequency patterns become rules; low-frequency patterns are exceptions on the same continuum. \\
\addlinespace
\textbf{Status of Grammar} & A pre-existing mental mechanism in the mind.  & A post-hoc descriptive abstraction created by linguists from observed patterns in texts.
\\
\addlinespace
\textbf{Validation Method} & Subjective sense of grammaticality; theoretical elegance; appeal to linguistic intuition and dogma. & ``Synthesis validates analysis.'' A theory is valid only if it can be used to reconstruct what it claims to explain. \\
\addlinespace
\textbf{View of Meaning} & Requires grounding in deep structure or real-world referents. Form and meaning are fundamentally separate. & Primarily relational, derived from network of connections between words. Most meanings exist within the axiomatic system of language. \\
\addlinespace
\textbf{Assessment of~LLMs} & Fundamentally flawed ``stochastic parrots'' lacking access to deep structure or grounded meaning. & 
A large-scale, empirical validation. Not broken in principle but incomplete in practice.
\\
\bottomrule
\end{tabularx}
\end{table*}

\begin{examplefloat*}[!ht]
  \centering
    \begin{exampleboxtwo}[title=Three studies from Latin's evolution into Romance languages]
    \textbf{(A) Analogical Regularization.}
    Classical Latin used a mix of rules to form words representing  quantities. While 17 was additive (\textit{septendecim} `seven-ten'), 18 and 19 were subtractive (\textit{duodeviginti} `two-from-twenty,' \textit{undeviginti} `one-from-twenty'). In the evolution of Latin to the Romance languages, these irregular subtractive forms were almost universally abandoned. These were replaced by more regular, compound forms analogically derived from simpler additive rules (e.g., Italian \textit{diciotto} `ten-and-eight').
    Additive rules served as the model for simplifying the rarer, more complex subtractive ones \cite{Manczak1958Tendances,Manczak1978Analogique,Manczak1996Analogicznego}.
    \vspace{5pt}\\ 
    \textbf{(B) Irregular Development due to Frequency.} The evolution of the Latin verb \textit{ambulare} `to walk' illustrates how usage frequency shapes language. The form of the word diverged based on its function.  Everyday use resulted in significant irregular shortening, leading to French \textit{aller}, Italian \textit{andare}, and Spanish \textit{andar}. Infrequent use in a specific context---e.g.,  `to amble,' which began as a specialized equestrian term---retained a more regular and archaic form: French \textit{ambler}, Italian \textit{ambiare}, and Spanish \textit{amblar}. The fact that high-frequency words are prone to irregular change, while less common words often preserve their regular, historical forms, is a common linguistic phenomenon \cite[\textit{inter alia}]{alma991018253099705524,Manczak1969a,Manczak1987,Manczak1988a,Manczak1996Nieregularny}. \vspace{5pt}\\
    \textbf{(C) Grammaticalization.}
    A fundamental shift in the history of Romance languages is the emergence of the `have' + participle construction to express the perfect tense (e.g., French \textit{il a chanté} `he has sung'). This developed from Classical Latin, where the verb \textit{habere} was a normal lexical verb meaning `to possess' or `to hold'. This new, constant grammatical use created a functional split. The word's original meaning has been mostly replaced by its modern role as a grammatical tense marker \cite{Manczak1996Semantyczny}. 
    \end{exampleboxtwo}
    \caption{Examples from the evolution of Latin into the Romance languages, demonstrating the impact of frequency and temporal character of grammar: (A) analogical regularization of numerals, (B) divergent evolution depending on usage frequency, and (C) grammaticalization from lexical verb to perfect tense marker.}    \label{ex:latin-evolution}
\end{examplefloat*}

\subsection{The Map Is Not the Territory}\label{sec:map}

Mańczak found it absurd that people would claim to study an  intangible ``system of signs'' or mental ``competence'' when their actual work involved poring over concrete texts. He proposed a simple, naturalistic definition: \sidenotetwo{Language Definition}\hlsnowtwo{language is the totality of all that is said and written.} 

The fundamental error of modern linguistics is the false equivalence between the product of a particular analysis and the object of the study itself \cite{critique}.

\paragraph{The Primacy of Frequency.} Where structuralists and their successors saw dichotomies, Mańczak saw a continuum governed by statistics. For example, in his view, the dichotomy between grammar and lexicon is false, because grammar is ``the quintessence, condensation, abbreviation, or generalization of lexicography'' \cite{Manczak1996Slownik}. Grammar covers only high-frequency patterns that apply to large classes of words. Lexicography, in contrast, handles the information for all individual words, including the less frequent and more idiosyncratic.

Even the distinction between ``rule'' and ``exception'' is quantitative, not qualitative: high-frequency patterns are rules, while low-frequency patterns are exceptions. We can interpret Mańczak's perspective in the light of Jaynes’ theory, according to which the absolute ``True/False'' statements of classic logic are seen as a special case of a more general, probabilistic logic \cite{jaynes2003probability}. In this framing, a Mańczakian ``rule'' corresponds to a high-plausibility inference, while an ``exception'' represents a low-plausibility inference.

\sidenote{Frequent Errors Become Norms}Importantly, according to \citet{critique} ``the rules of grammar, once abstracted from texts, do not prevent the subsequent evolution of language. This evolution consists of \hlsnow{errors which, if their frequency increases sufficiently, become new norms}. Conversely, old norms, if their frequency diminishes, become errors.''

Example~\ref{ex:latin-evolution} presents three examples from the evolution of Latin into Romance languages that illustrate how frequency drives linguistic change and, crucially, how grammatical rules themselves are merely abstractions from this ongoing evolution.

\paragraph{Language Acquisition.} Mańczak's claim that frequency of use in language is a primary organizing principle is strongly supported by decades of research in cognitive science. Studies show that frequency affects nearly every level of language processing in humans, from the recognition of sounds and words to the processing of syntax \cite[\textit{inter alia}]{SAFFRAN1996606,Romberg2010,10.7554/eLife.101802}. High-frequency patterns are processed more quickly and accurately, and learned earlier by children, confirming that the human mind is finely tuned to the frequency of exposure to particular language elements \cite{Ellis_2002}.

\paragraph{} This statistics-based view of language contrasts strikingly with dominant linguistic paradigms. As Mańczak noted, ``the fact that, in the three hundred pages of [de Saussure's] \textit{Course in General Linguistics} the term \textit{frequency of use} does not appear even once, has serious implications'' \cite{critique}. \sidenotethree{Result of Ignoring Frequency} \hlsnowthree{When frequency is ignored, linguists must invent increasingly elaborate theoretical constructs}---deep structures, innate faculties, universal grammars---to explain what simple statistical patterns readily predict.

\begin{examplefloat}[b]
  \centering
    \begin{examplebox}[title=Validation through synthesis]
    \textbf{The analysis of a simple sentence.}
    Mańczak examined how linguists with competing methodologies analyze a simple English sentence: \textit{Sincerity may frighten the boy}. He observed that creating such a sentence using generative grammar required a long, convoluted process.
    \vspace{5pt}\\
    In contrast, to synthesize the sentence, one needs only a few simple rules: ``1) Every noun can be accompanied by a third‑person verb that agrees with it in number.
    2) Certain verbs, including \textit{may}, can be accompanied by infinitives without the particle \textit{to}. Since these verbs are rare, grammars enumerate them.
    3) Transitive verbs (including \textit{frighten}) are accompanied by a noun. Since these verbs are common, grammars refer to dictionaries for detailed information. 4) Every noun can be accompanied by an article under certain conditions. Grammars enumerate these conditions.
    5) The article precedes the noun, the infinitive follows the verb it accompanies, etc.'' \cite{Manczak1996GramatykaTransformacyjno}.
    \vspace{5pt}\\
    These straightforward rules highlight the gap between the theoretical complexity of Chomsky's analysis and the practical utility of Mańczak's alternative approach. 
    
    \end{examplebox}
    \caption{The theory requiring immense abstract machinery to analyze a simple sentence that can be generated with a few practical rules has failed the fundamental test of synthesis.}
  \label{ex:sentence-analysis}
\end{examplefloat}

\subsection{Talkin' All About That Syntax When You Can't Even Generate a Haiku}

Once invented, these constructs then take on a life of their own. Critics of LLMs now assert that \textit{true} linguistic competence requires an innate ``deep structure'' or ``internal form of language.'' This claim mistakenly interprets the linguist's abstract analysis as the functional prerequisite of language itself. As \citet{Manczak1996Gramatyka} puts it:
\begin{quote}
\semismall
If a layman were to try to convince a chemist that water is composed not of two elements (i.e., oxygen and hydrogen) but of four, the chemist would shrug his shoulders and conclude that the third and fourth elements exist only in the layman's imagination, because chemists have succeeded countless times in synthesizing water from only two elements.
Now, the principle that synthesis [reconstructing a coherent whole] is required to validate analysis [breaking things down into components] \ldots{} should be applied to linguistics.
Linguists who analyze text should look for only those components that are essential for its synthesis and must validate their analyses by means of synthesis.
\end{quote}

He similarly criticized proponents of the generative and transformational grammars that dominated 20th-century linguistics. Mańczak noted with irony that while champions of these theories filled entire volumes with abstract formalisms, their analysis was never validated by synthesis. More than half a century after the ``generative'' turn, its adherents had ``not yet written a single generative or transformational grammar of a concrete language'' \cite{Manczak1996GramatykaTransformacyjno}. The roar of generative theory had produced not even a whisper of practical synthesis.

\sidenotethree{Failure to Synthesize} This failure becomes evident if we review the actual analyses. As Mańczak pointed out, \hlsnowthree{``Chomsky, to analyze a sentence as simple as \textit{sincerity may frighten the boy}, needed 10 pages in his book} \textit{Aspects of the Theory of Syntax}, whereas to reconstruct this sentence, it is enough to cite only five simple positional rules'' as illustrated in Example~\ref{ex:sentence-analysis} \cite{Manczak1996GramatykaTransformacyjno}. The principle that should guide any scientific grammar---``synthesis validates analysis''---was, as Mańczak observed, ``completely unknown to Chomsky.''

Yet linguists tolerated lack of validation for decades, mistaking Chomsky's theoretical complexity for demonstrated accuracy.
At the same time, when they needed to assess grammaticality, they searched corpora or polled native speakers instead of consulting a generative rule system.
Today, practitioners of linguistics already operate in Mańczak's world, even if linguistic theorists haven't caught up.

Stunningly effective LLMs have arrived as the answer to the unfulfilled promise of generative grammar.
Their capacity to synthesize coherent language based on probabilistic analysis of text rather than abstract rules or postulated deep structures is the ultimate vindication of Mańczak's approach. 

Empirically,\sidenotetwo{LLMs and Frequency} LLM performance increases smoothly with the amount of pretraining data \citep{kaplan2020scalinglawsneurallanguage,hoffmann2022trainingcomputeoptimallargelanguage}. Pretraining minimizes expected next‑token surprisal (\textit{cross‑entropy}), pushing the model's conditional predictions to match empirical next‑token frequencies. \hlsnowtwo{Estimation of the language's frequency structure} improves and sharpens with a larger training set, especially in the long tail. Interestingly, models naturally retain the useful token frequency information in embeddings \citep{zhou2021frequencybaseddistortionscontextualizedword,DBLP:journals/corr/abs-1809-06858}. 

LLMs' success is not a mystery or a ``stochastic parrot'' trick. It is a large-scale validation of Mańczak's central thesis: language is text, and frequency is not a secondary, peripheral aspect, but its primary organizing force.\footnote{For a complementary view, see \citet{piantadosi2023modern}, who argues that LLMs refute Chomsky's theory.} 

\subsection{An Obituary for the Language Organ}

According to the Chomskyan paradigm, language acquisition requires a specialized ``language organ'' because children lack sufficient linguistic input to develop language from pattern recognition alone.
Mańczak rejected this view, arguing that  debates about hypothetical brain  structures fell outside the proper naturalistic focus of linguistics \cite{Grochowski_2017,critique}.

We see now that his skepticism was justified.
The Chomskyan paradigm has been challenged by  evidence from cross-linguistic research and developmental psychology, causing many experts to abandon it \cite[\textit{inter alia}]{Ibbotson2016,Pullum2002,Christiansen2008,mothers,tomasello,Bybee2006}. Instead, a ``usage-based'' alternative has emerged. Its proponents argue that children learn grammar from the ground up by applying general cognitive tools like pattern recognition and categorization to the sounds they hear.
\begin{table*}[t!]
\centering
\semismall
\caption{Architectural evolution and its impact on analogical capability.}
\label{tab:architecture_analogy}
\begin{tabularx}{\textwidth}{l p{3.9cm} p{4.7cm} p{4.75cm}}
\toprule
\textbf{Model} & \textbf{Core Mechanism} & \textbf{Representation of Words} & \textbf{Analogical Capability} \\
\midrule
\textbf{N-gram} & Surface-level counting of token co-occurrence; $P($word $|$ previous $n-1$ words$)$. & Atomic, discrete symbols. No notion of similarity between words. Sentences ``Anna likes cats'' and ``Lily loves dogs'' are not similar. & \textbf{None.} Can reproduce frequent sequences but cannot grasp words or sentences similarity. \\
\addlinespace
\textbf{CBOW} & Learning vector representation of meaning by predicting a word from its context. & Dense, low-dimensional vectors. Words occurring in similar contexts are mapped to nearby points in a geometric space. & \textbf{External to model.} Series of analogous word pairs can be demonstrated to share similar geometric relationship. \\
\addlinespace
\textbf{LLMs} & Learning relationships between learned vector representations of meaning. 
& Dense, high-dimensional vectors that are dynamically adjusted based on the specific context of the entire sequence. & \textbf{Inherent, core mechanism.} The model operates on a series of embeddings and generalizes patterns. \\
\bottomrule
\end{tabularx}
\end{table*}
This evidence-based theory sees grammar as an emergent property of history and psychology---a set of templates discovered by observing that some sentences or word forms are built analogically. It aligns perfectly with Mańczak's insistence on an empirical, text-first science of language.\footnote{See \citet{jbp:/content/journals/10.1075/cf.23017.gol} for a usage‑based constructionist account explicitly linking constructions to LLMs.}

To criticize the ability of LLMs to generate plausible human speech as a mere application of learned patterns is to miss the point that use of analogy is an essential aspect of linguistic competence.
LLMs can generate plausible real-time answers in part because they recognize---during previous training---the same frequency patterns that created human grammar in the first place.

The\sidenotethree{Key Leap: \emph{Relations} Between Embeddings} success of modern LLMs stems from a crucial architectural innovation: the replacement of flat, surface-level n-gram counts with high-dimensional embeddings (Table~\ref{tab:architecture_analogy}). 
Earlier n-gram models, limited by their reliance on simple tables of memorized word combinations, failed to recognize that the sentences `Anna likes cats' and `Lily loves dogs' are analogous. While CBOW models could demonstrate that analogous word pairs share a similar geometric relationship, the breakthrough improvement was to \hlsnowthree{operate on sequences of learned vector representations.}
When presented with a novel problem to solve, Transformer uses its vast internalized map of relationships to find and apply the closest learned analogy.

This ability to represent and manipulate relationships---the very essence of analogy---is the key to genuine linguistic generalization.

\subsection{This Page is Intentionally Left Ungrounded}
\label{sec:meaning_axioms}

Perhaps the most persistent criticism of LLMs is that they are ungrounded: they manipulate symbols (\textit{form}) without  access to their real-world referents (\textit{meaning}).\footnote{See \citet{bender-koller-2020-climbing} for a representative formulation of the form-vs-meaning objection.} \sidenotetwo{Axiomatic Meaning}While Mańczak did not directly engage with modern semantic theories, his view  aligns with a core assumption of componential and reductionist semantics: the \hlsnowtwo{necessity of undefinable primitives} \cite{Grochowski_2017}.

Mańczak argued that attempting to describe language without external reference  (ignoring meaning \textit{entirely}) was a descent into a ``nebulous darkness.'' He proposed a simple resolution \cite{Manczak1996Slownik}:
\begin{quote}
\semismall
Just as in mathematics, most but not all statements can be proved (with the help of other statements) \ldots{} most words in a given language can be defined with the help of other words, with the unavoidable exception that [to avoid circularity] the meaning of certain words must be taken as self-evident (axiomatic).
\end{quote}

Nevertheless, he recognized that the meaning of most words is relational, derived from an intricate web of connections between terms.\footnote{Concepts without referents---such as ``perpetual motion machine'' or ``king of San Francisco''---further illustrate how meaning can exist purely through relational networks \cite{piantadosi2022meaningreferencelargelanguage}.}

An LLM correctly using a concept like ``justice'' requires only mastery of the vast, multidimensional web of relationships that connect that word to ``fairness,'' ``law,'' ``equality,'' ``crime,'' and thousands of other terms. Whether this represents a human-like \textit{understanding} is irrelevant.

We do not demand that a calculator understands what ``1+2'' \textit{truly means} to accept its utility. We do not dismiss the results of theorem-proving software because it cannot understand the philosophical basis of Zermelo-Fraenkel set theory. While simple counting may be grounded in early childhood experiences, advanced mathematical reasoning involves manipulation of a highly formal system according to a set of rules. For most people, including professional mathematicians, the ``meaning'' of an axiom lies in its role within that formal system, rather than an intuitive or sensory experience.

Demanding that a language model meet a higher standard of ``grounded meaning'' is misguided.
In the Mańczakian view, the relevant test for LLM quality is not whether the model has access to an outside world, but whether it has mastered the internal, relational logic of the textual world it was given.\footnote{This textual world might suffice. \citet{mandelkern-linzen-2024-language} argue that LLMs can use words to talk about real things because training texts already link those words to the physical world. In their view, the key issue is whether the model counts as part of our speech community.}

\section{Conclusion}

LLM users often feel amazed, disappointed, or a mixture of both. They're amazed that LLMs can produce human-like text about any topic. They're disappointed when models confidently present factually incorrect information. 

Beyond these reactions, prominent linguists argue that LLMs are intrinsically flawed because of gaps in  linguistic competence and inability to ``understand'' language.
They call for models to demonstrate knowledge of ``deep structure'' or reliance on ``grounded meaning''---concepts their theories treat as requirements of language.

\sidenote{LLMs\\Merely\\Model Language} We argue that by applying these theory-laden standards, such critics interfere with useful analysis.
\hlsnow{The ``stochastic parrots'' do not merely mimic language but in fact reveal what language has been all along.}
LLMs are imperfect tools not because they \textit{fail} to model language but because they \textit{only} model language.

The implications of our proposed shift in perspective extend beyond theoretical linguistics.
The path to improving \textit{linguistic} competence of LLMs is to design, evaluate, and deploy systems that have already proven mastery of language's relational logic. Satisfying abstract theoretical requirements for language competence---demanded by the vocal critics of LLMs---will neither improve linguistic competence nor bring us closer to Artificial General Intelligence.

Mańczak saw it clearly: language is the totality of texts, and frequency is its organizing principle. Decades later, engineers unknowingly built LLMs on this very foundation. Their models now draft our contracts, structure our arguments, and increasingly impact our futures---making Mańczak's case more powerfully than any argument could.

\begin{figure}[b]
\begin{translationtwo}
\subsection*{Witold Mańczak (1924–2016)}

Witold Mańczak was a Polish linguist specializing in Indo-European, Romance, and Slavic studies. Over his career, he published 24 books and more than 960 other texts, developing a distinct theoretical framework for language analysis  \cite{Debowiak2014,Debowiak_Przemyslaw_Witold_2016,Grochowski_2017}.
\vspace{5pt}\\
A defining feature of Mańczak's work was his application of statistical methods to verify linguistic hypotheses. He argued that linguistic analysis should be based on quantifiable data, and proposed that language development is driven by three factors: regular phonetic development, analogical development, and irregular phonetic development caused by frequency of use. With his data-centric approach, Mańczak challenged several foundational linguistic theories (see Appendix~\ref{appendix:foreword}).  
\vspace{5pt}\\
His life's work stands as a testament to a tireless pursuit of verifiable statements, establishing him as the creator of a coherent, statistics-based theory of language and one of the most significant linguists of his era.\end{translationtwo}\end{figure}

\section{Limitations}

We acknowledge that proposing a novel theoretical lens for understanding language and LLMs leaves some critical questions unaddressed.

\paragraph{Your paper ignores urgent ethical dilemmas.} Carefully defining a technology is a prerequisite for ethical debate. We address the former to set the stage for the latter. 

\paragraph{A text-only definition of language excludes some linguistic subfields, such as pragmatics, psycholinguistics, etc.}
To qualify as linguistic, a claim must generate falsifiable predictions about \textit{distributions over texts} or about \textit{observable properties of utterances}, all of which must be testable by corpus statistics or experiments. Pragmatics satisfies these conditions when it predicts context‑conditioned choices among forms (e.g., hedging, politeness markers). Sociolinguistics does so when it predicts socially conditioned alternations in variants and styles across communities. Ethnolinguistics is linguistic when it predicts culturally conditioned textual patterns in a community corpus (e.g., motifs, metaphor families, collocational frames).

Claims that fail to meet these conditions may still be valuable—in sociology, psychology, or cultural studies—but they do not count as linguistic in the Mańczakian sense.

\paragraph{Your focus on relational meaning seems to embrace a purely structuralist view of semantics.}
We do not claim to offer a complete theory of meaning. Our central thesis is that the vast majority of the time ``meaning'' can be inferred (and in the case of LLMs is inferred) solely from the relational structure of the text. The success of LLMs demonstrates how much linguistic competence can be achieved by assigning a central role to distributional semantics, and a limited role---if any---to direct grounding.

\paragraph{LLMs are trained on unrepresentative corpora, very distant from ``the totality of all that is said and written.''} This is true and aligns with our view that LLMs are not fundamentally flawed, but incomplete in practice. The Mańczakian framework offers a clear path forward, favoring ``rational selection of texts'' based on circulation and influence \cite{Manczak1996Racjonalny,Manczak1961}. A truly Mańczakian approach is a principled, frequency-weighted corpus construction that reflects how language is actually used.

\paragraph{LLMs lack a ``world model.''} The Mańczakian framework we adopt posits that language is not a map of the physical world but a self-contained universe of texts. From this perspective, the demand for grounding in physical reality is misplaced. Our model should be seen not as a flawed attempt to simulate a mind interacting with a physical world but as a successful and direct model of language itself.

\paragraph{You are closing off valuable inquiry into the cognitive plausibility of LLMs.}
Our goal is not to declare cognitive comparisons invalid but to argue that they should not be the \textit{primary} benchmark.
An LLM is, first and foremost, a direct model of textual corpus. This is not a metaphor; it is a description of its construction and function.
We argue that any comparison to human cognition must come \textit{after} this primary, non-metaphorical evaluation is established. The Mańczakian framework provides a necessary baseline. 

\paragraph{Many of Mańczak's core ideas are central to the well-established ``usage-based linguistics.''}
While Mańczak’s ideas do align with modern usage-based linguistics, he reached similar conclusions decades earlier through the empirical analysis of contemporary and historical texts.

First, Mańczak's radical, text-only simplicity provides a direct rebuttal to critiques of LLMs grounded in Saussurean or generative theories: his framework simply discards abstractions that cannot be found in the textual record.
Second, his focus on the structure of the input text itself---rather than human cognitive processing---aligns directly with how text-trained models actually function.

The fact that Mańczak's textual analysis independently yielded insights that are now central to cognitivism makes his framework particularly compelling.

\paragraph{If linguistic competence requires merely applying frequency-weighted patterns, how do you explain creativity?}
High-frequency patterns don't mechanically reproduce themselves; they serve as templates for novel combinations.
LLMs demonstrate this empirically. Creativity isn't the opposite of pattern utilization---it's pattern mastery.

\section*{Acknowledgments}

I am grateful to Geoffrey Laff for his insightful comments and suggestions, which significantly improved the style and clarity of this paper.

I also thank Daniel Dzienisiewicz, Filip Graliński, Michał Pietruszka, Michał Turski, and Jordy Van Landeghem for reading a draft and offering valuable remarks.


\clearpage
\appendix

\section{Selected Contributions to Historical and Comparative Linguistics}\label{appendix:foreword}

\begin{translation}
This summary was initially published by Mańczak in French and served as a foreword to \textit{Linguistique générale et linguistique indo-européenne}. It outlines his career-long pursuit against fundamental methodological flaws of modern linguistics. Some of these findings oppose prevailing views---and that is precisely why his methodological proposal cannot be ignored.
I strongly encourage the reader to seek out the original publications.
\end{translation}

\subsection*{English Translation}

The fundamental problem of linguistics is that of the criteria of truth. Unfortunately, this problem is taboo. Given that linguistics has existed for two thousand years and that the \textit{Linguistic Bibliography} recorded 21,000 works for the year 2001, it follows that linguists have published, in total, several hundreds of thousands of works, and yet none of these has been devoted to the criteria of truth. Even the term \enquote{criteria of truth} is never used by linguists. This is an extraordinary thing, considering that linguists unanimously agree that linguistics is a science, and that science is a search for truth. Why do linguists keep the question of distinguishing true from false in their discipline secret?

This enigma has intrigued me for a long time. As linguistic works provide no information capable of resolving this question, I began to observe how linguists react when they learn an opinion previously unknown. To my great astonishment, I found that linguists never intend to verify the opinion in question, but are interested only in the question of who shares this opinion. If they learn that this opinion is shared by one or more authorities, they consider it true. If, on the contrary, they learn that this opinion comes from someone who does not have the reputation of being an authority, this view appears false to them. The criterion of truth used by linguists is the following: X has formulated an opinion, X is an authority, therefore this opinion is true; Y has formulated an opinion, Y is not an authority, consequently this opinion is false. Obviously, this criterion of truth is medieval and unscientific, which is why linguists prefer not to talk about it.

In this state of affairs, I reflected on the criteria of truth likely to be employed in linguistics. I came to the conclusion that linguists can resort to statistics (and, exceptionally, to experiment) and that, in the science of language, many opinions rely on faith in the infallibility of authorities and are invalidated by statistical data. Here are some examples:

\begin{enumerate}[wide, labelwidth=!, labelindent=0pt]
\item In all languages, the form of words depends on three main factors, not only on regular phonetic development and analogical development, but also on what I call irregular phonetic development due to frequency (\textit{Le développement phonétique des langues romanes et la fréquence}, Kraków, 1969; \textit{Słowiańska fonetyka historyczna a frekwencja}, Kraków, 1977; \textit{Frequenzbedingter unregelmäßiger Lautwandel in den germanischen Sprachen}, Wrocław, 1987). According to a professor of applied mathematics, the chance that the theory of irregular phonetic development due to frequency is erroneous is less than 1 in 10 million (\textit{Etymologia przyimka dla a nieregularny rozwój fonetyczny spowodowany frekwencją}, Prace Filologiczne 60, 2011, p.~189--195).

\item Bartoli's \enquote{norm} according to which lateral areas are more archaic than central areas is invalidated by statistical data (\textit{La Roumanie et l'Espagne sont-elles des territoires archaïques de la Romania?}, Limba românǎ, limbǎ romanicǎ. Omagiu acad. M. Sala la împlinirea a 75 de ani, Bucureşti, p.~313--317).

\item Since 1925, when Meillet introduced the notion of \enquote{empty slot} (\textit{case vide}), it has been imagined that phonetic evolution consists of filling \enquote{empty slots} in phonological systems. But I have examined a large number of facts and have come to the conclusion that it is not symmetry, but asymmetry that characterizes languages, that it is possible to formulate a law according to which more frequently used linguistic elements are more differentiated than less used elements (\textit{Do the ``cases vides'' exist?}, Linguistique générale et linguistique indo-européenne, Kraków, 2008, p.~59--62).

\item Laryngeal theory is invalidated by statistical data (\textit{Critique de la théorie des laryngales}, Analecta Indoeuropaea Cracoviensia I. Safarewicz memoriae dicata, Cracoviae, 1995, p.~237--247; \textit{Encore un argument contre la théorie des laryngales}, Lingua Posnaniensis 46, 2004, p.~41--44).

\item In my opinion, Verner's law requires revision (\textit{La restriction de la règle de Verner à la position médiane et le sort du s final en germanique}, Historische Sprachforschung 103, 1990, p.~92--101; \textit{La règle de Verner s'applique-t-elle à la position finale?}, Historische Sprachforschung 109, 1996, p.~110--116).

\item In light of statistical data, Old Church Slavonic is a compromise between the Macedono-Bulgarian dialect and the Moravo-Pannonian speech (\textit{Przedhistoryczne migracje Słowian i pochodzenie języka staro-cerkiewno-słowiańskiego}, Kraków, 2004; \textit{Pochodzenie języka staro-cerkiewno-słowiańskiego a Kodeks zografski}, Warszawa, 2006).

\item In light of statistical data, the original homeland of the Indo-Europeans is identical with that of the Slavs (\textit{De la préhistoire des peuples indo-européens}, Kraków, 1992; \textit{L'habitat primitif des Indo-Européens se trouvait-il vraiment en Arménie?}, Folia Orientalia 33, 1997, p.~65--74).

\item The German orientalist Ludolf (17th c.) was the first to affirm that \enquote{die Sprachverwandtschaft offenbart sich nicht im Wörterbuch, sondern in der Grammatik} [language relationship reveals itself not in the dictionary, but in grammar]. But one can justify the division of Indo-European languages into Germanic, Slavic, Baltic, Romance, etc. only through lexical convergences, and not inflectional or phonetic ones (\textit{La classification des langues romanes}, Kraków, 1991, p.~22--36).

\item The number one problem of Romance etymology is that of verbs meaning \enquote{to go}: Fr. \textit{aller}, It. \textit{andare}, Sp. \textit{andar}, Prov. \textit{ana}, etc. Since the 16th century, about sixty etymologies have been proposed in total, which is a record, and not only for Romance etymology. Among researchers, there are adherents to monogenesis (affirming that all these forms come from, for example, \textit{ambulare}) and advocates of polygenesis (claiming, for example, that \textit{aller} < *\textit{advehulare}, \textit{andar} < *\textit{am(bi)vehitare}, \textit{ana} < *\textit{amvehinare}). Probability calculus allows this question to be decided in favor of monogenesis (\textit{Une étymologie romane controversée: aller, andar, etc.}, Revue roumaine de linguistique 19, 1974, p.~89--101; \textit{Étymologie de fr. aller, esp. andar, etc. et calcul des probabilités}, Revue roumaine de linguistique 20, 1975, p.~735--739).

\item Since 1435, it has been affirmed that Romance languages derive from Vulgar Latin, but, in light of statistical data, they derive from Classical Latin (\textit{Le problème de l'origine des langues romanes dans le livre de H. Lüdtke et celui de R. Kiesler}, Actes du XXVe Congrès International de Linguistique et de Philologie Romanes, t. VI, Berlin, 2010, p.~207--211).

\item Since Jordanes, that is, for 1400 years, it has been estimated that the original homeland of the Goths was in Scandinavia. But comparison of parallel texts in Gothic, High German, Middle German, Low German, Danish and Swedish revealed that the original homeland of the Goths was in the southernmost part of ancient Germania (\textit{Le mythe de l'origine scandinave des Goths}, L'art de la philologie. Mélanges en l'honneur de L. Löfstedt, Helsinki, 2007, p.~137--145).

\item The division of words into stressed and unstressed (articles, pronouns, prepositions, etc.), which dates back to Antiquity, is the result of a false generalization. It is true that there are homonymies \textit{le vent} = \textit{levant}, \textit{à voir} = \textit{avoir}, and \textit{moi} = \textit{émoi} and that the syllables \textit{le-}, \textit{a-}, \textit{é-} in \textit{levant}, \textit{avoir}, \textit{émoi} are unstressed, but it is erroneous to conclude from this that \textit{le}, \textit{à}, and are unstressed because \enquote{stressed} words are treated in the same way. \textit{Dix vers}, \textit{vingt cœurs}, \textit{va tôt}, pronounced without pauses, are homonymous with \textit{divers}, \textit{vainqueur}, \textit{Watteau}, where the syllables \textit{di-}, \textit{vain-}, \textit{Wa-} are unstressed. It is affirmed that \textit{Long vient} = stressed word + stressed word, while \textit{l'on vient} = proclitic + stressed word, but a very simple experiment proves that these expressions are homonymous (\textit{La division des mots en toniques et atones est-elle justifiée?}, Lingua Posnaniensis 32--33, 1991, p.~181--185).

\item Since Antiquity, the question of what constitutes the difference between proper nouns and common nouns has been discussed. About ten definitions of the proper noun have been proposed so far, none of which applies to all proper nouns. In my opinion, the difference between proper nouns and common nouns consists in the fact that common nouns are, in the vast majority of cases, translated from one language to another, while proper nouns almost never are. For example, a common noun like \textit{ville} is translated into Italian as \textit{città}, into English as \textit{town}, etc., whereas a proper noun like \textit{Paris} is not, cf. It. \textit{Parigi}, Eng. \textit{Paris}, etc. Among all definitions of the proper noun, mine suffers the fewest exceptions (\textit{La notion de nom propre}, Proceedings of 13th International Congress of Onomastic Sciences, Kraków, 1982, p.~101--106).
\end{enumerate}

\subsection*{French Original}

Le problème fondamental de la linguistique est celui des critères de vérité. Malheureusement, cette question constitue un tabou. Étant donné que la linguistique existe depuis deux mille ans et que la Bibliographie linguistique a enregistré, pour l’année 2001, 21 000 travaux, il en résulte que les linguistes en ont publié, au total, plusieurs centaines de milliers, et pourtant aucun de ces derniers n’a été consacré aux critères de vérité. Même le terme « critères de vérité » n’est jamais employé par les linguistes. C’est une chose extrêmement étrange, si l’on considère que les linguistes sont unanimes pour dire que la linguistique est une science, et que la science n’est pas autre chose qu’une recherche de la vérité. Pourquoi donc les linguistes gardent-ils un secret sur la question de savoir comment ils distinguent le vrai du faux dans leur discipline?

Cette énigme m’a intrigué depuis longtemps. Comme les travaux linguistiques ne fournissent aucun renseignement susceptible de résoudre cette question, j’ai commencé à observer comment les linguistes réagissent quand ils apprennent une opinion qui leur était inconnue auparavent. A mon grand étonnement, j’ai constaté que les linguistes n’ont jamais l’intention de vérifier l’opinion en question, mais s’intéressent uniquement à la question de savoir qui partage cette opinion. S’ils apprennent que cette opinion est partagée par une ou plusieurs autorités, ils considèrent cette opinion comme vraie. Si, au contraire, ils apprennent que cette opinion provient de quelqu’un qui n’a pas la réputation d’être une autorité, cette vue leur paraît fausse. Il en résulte que le critère de vérité utilisé par les linguistes est le suivant: X a formulé une opinion, X est une autorité, par conséquent cette opinion est vraie; Y a formulé une opinion, Y n’est pas une autorité, par conséquent cette opinion est fausse. Évidemment, ce critère de vérité est médiéval, non scientifique, et c’est la raison pour laquelle les linguistes préfèrent ne pas en parler.

Dans cet état de choses, il m’est venu à l’esprit de réfléchir sur les critères de vérité susceptibles d’être employés en linguistique et je suis arrivé à la conclusion que les linguistes peuvent recourir à la statistique (et, exceptionnellement, à l’expérience) et que, dans la science du langage, il y a beaucoup d’opinions qui s’appuient sur la foi en l’infaillibilité des autorités et qui sont infirmées par des données statistiques. Voici quelques exemples.

\begin{enumerate}[wide, labelwidth=!, labelindent=0pt]
\item Dans toutes les langues, la forme des mots dépend de trois facteurs principaux, non seulement du développement phonétique régulier et du développement analogique, mais aussi de ce que j’appelle un développement phonétique irrégulier dû à la fréquence (\textit{Le développement phonétique des langues romanes et la fréquence}, Kraków, 1969; \textit{Słowiańska fonetyka historyczna a frekwencja}, Kraków, 1977; \textit{Frequenzbedingter unregelmäßiger Lautwandel in den germanischen Sprachen}, Wrocław, 1987). De l’avis d’un professeur de mathématiques appliquées, la chance que la théorie du développement phonétique irrégulier dû à la fréquence soit erronée, est moindre que 1 sur 10 millions (\textit{Etymologia przyimka dla a nieregularny rozwój fonetyczny spowodowany frekwencją}, Prace Filologiczne 60, 2011, p. 189–195).

\item La « norme » de Bartoli d’après laquelle les aires latérales sont plus archaïques que les aires centrales, est infirmée par des données statistiques (\textit{La Roumanie et l’Espagne sont-elles des territoires archaïques de la Romania?}, Limba românǎ, limbǎ romanicǎ. Omagiu acad. M. Sala la împlinirea a 75 de ani, Bucureşti, p. 313–317).

\item Depuis 1925, où Meillet a introduit la notion de « case vide », on imagine que l’évolution phonétique consiste à remplir des « cases vides » dans les systèmes phonologiques. Mais j’ai examiné un grand nombre de faits et suis arrivé à la conclusion que ce n’est pas la symétrie, mais l’asymétrie qui caractérise les langues, qu’il est possible de formuler une loi d’après laquelle les éléments linguistiques plus employés sont plus différenciés que les éléments moins utilisés (\textit{Do the “cases vides” exist?}, Linguistique générale et linguistique indo-européenne, Kraków, 2008, p. 59–62).

\item La théorie des laryngales est infirmée par des données statistiques (\textit{Critique de la théorie des laryngales}, Analecta Indoeuropaea Cracoviensia I. Safarewicz memoriae dicata, Cracoviae, 1995, p. 237–247; \textit{Encore un argument contre la théorie des laryngales}, Lingua Posnaniensis 46, 2004, p. 41–44).

\item A mon avis, la règle de Verner exige une révision (\textit{La restriction de la règle de Verner à la position médiane et le sort du s final en germanique}, Historische Sprachforschung 103, 1990, p. 92–101; \textit{La règle de Verner s’applique-t-elle à la position finale?}, Historische Sprachforschung 109, 1996, p. 110–116).

\item A la lumière de données statistiques, le vieux slave est un compromis entre le dialecte macédo-bulgare et le parler moravo-pannonien (\textit{Przedhistoryczne migracje Słowian i pochodzenie języka staro-cerkiewno-słowiańskiego}, Kraków, 2004; \textit{Pochodzenie języka staro-cerkiewno-słowiańskiego a Kodeks zografski}, Warszawa, 2006).

\item A la lumière de données statistiques, l’habitat primitif des Indo-Européens est identique avec celui des Slaves (\textit{De la préhistoire des peuples indo-européens}, Kraków, 1992; \textit{L’habitat primitif des Indo-Européens se trouvait-il vraiment en Arménie?}; Folia Orientalia 33, 1997, p. 65–74).

\item L’orientaliste allemand Ludolf (17e s.) a été le premier à affirmer que « die Sprachverwandtschaft offenbart sich nicht im Wörterbuch, sondern in der Grammatik ». Mais on peut justifier la division des langues indo-européennes en germaniques, slaves, baltes, romanes, etc. uniquement par des convergences lexicales, et non flexionnelles ou phonétiques (\textit{La classification des langues romanes}, Kraków, 1991, p. 22–36).

\item Le problème numéro un de l’étymologie romane est celui des verbes ayant pour sens « aller » : fr. \textit{aller}, it. \textit{andare}, esp. \textit{andar}, prov. \textit{ana}, etc. Depuis le 16e siècle, on a, au total, proposé une soixantaine d’étymologies, ce qui est un record, et cela non seulement pour l’étymologie romane. Parmi les chercheurs, il y a des adhérents à la monogenèse (affirmant que toutes ces formes proviennent, par exemple, de \textit{ambulare}) et des adeptes de la polygenèse (prétendant, par exemple, que \textit{aller < *advehulare, andar < *am(bi)vehitare, ana < *amvehinare}). Le calcul des probabilités permet de trancher cette question en faveur de la monogenèse (\textit{Une étymologie romane controversée: aller, andar, etc.}, Revue roumaine de linguistique 19, 1974, p. 89–101 ; \textit{Étymologie de fr. aller, esp. andar, etc. et calcul des probabilités}, Revue roumaine de linguistique 20, 1975, p. 735–739).

\item Depuis 1435, on affirme que les langues romanes proviennent du latin vulgaire, mais, à la lumière de données statistiques, elles sont issues du latin classique (\textit{Le problème de l’origine des langues romanes dans le livre de H. Lüdtke et celui de R. Kiesler}, Actes du XXVe Congrès International de Linguistique et de Philologie Romanes, t. VI, Berlin, 2010, p. 207–211).

\item Depuis Jordanès, c’est-à-dire depuis 1400 ans, on estime que l’habitat primitif des Goths se trouvait en Scandinavie. Mais la comparaison de textes parallèles en gotique, allemand supérieur, moyen allemand, bas allemand, danois et suédois a révélé que l’habitat primitif des Goths se trouvait dans la partie la plus méridionale de la Germanie ancienne (\textit{Le mythe de l’origine scandinave des Goths}, L’art de la philologie. Mélanges en l’honneur de L. Löfstedt, Helsinki, 2007, p. 137–145).

\item La division des mots en toniques et atones (articles, pronoms, prépositions, etc.), qui remonte à l’Antiquité, est le résultat d’une fausse généralisation. Il est vrai qu’il y a des homonymies \textit{le vent = levant}, \textit{à voir = avoir}, \textit{et moi = émoi} et que les syllabes \textit{le-, a-, é-} dans \textit{levant, avoir, émoi} sont atones, mais il est erroné d’en conclure que le, à, et sont atones parce que les mots « toniques » sont traités de la même manière. \textit{Dix vers, vingt cœurs, va tôt}, prononcés sans pauses, sont homonymes de \textit{divers, vainqueur, Watteau}, où les syllabes \textit{di-, vain-, Wa-} sont atones. On affirme que \textit{Long vient} = mot tonique + mot tonique, alors que \textit{l’on vient} = proclitique + mot tonique, mais une expérience bien simple prouve que ces expressions sont homonymes (\textit{La division des mots en toniques et atones est-elle justifiée?}, Lingua Posnaniensis 32–33, 1991, p. 181–185).

\item Depuis l’Antiquité, on discute la question de savoir en quoi consiste la différence entre noms propres et noms communs. On a jusqu’ici proposé une dizaine de définitions du nom propre, dont aucune ne s’applique à tous les noms propres. A mon avis, la différence entre noms propres et noms communs consiste en ce que les noms communs sont, dans la grande majorité des cas, traduits d’une langue à l’autre, tandis que les noms propres ne le sont presque jamais. Par exemple, un nom commun comme \textit{ville} est traduit en italien par \textit{città}, en anglais par \textit{town}, etc., alors qu’un nom propre comme \textit{Paris} ne l’est pas, cf. it. \textit{Parigi}, angl. \textit{Paris}, etc. Parmi toutes les définitions du nom propre, la mienne souffre le moins d’exceptions (\textit{La notion de nom propre}, Proceedings of 13th International Congress of Onomastic Sciences, Kraków, 1982, p. 101–106).
\end{enumerate}

\clearpage

\bibliography{custom}

\begin{thebibliography}{46}
\providecommand{\natexlab}[1]{#1}

\bibitem[{Bender et~al.(2021)Bender, Gebru, McMillan-Major, and Shmitchell}]{10.1145/3442188.3445922}
Emily~M. Bender, Timnit Gebru, Angelina McMillan-Major, and Shmargaret Shmitchell. 2021.
\newblock \href {https://doi.org/10.1145/3442188.3445922} {On the dangers of stochastic parrots: Can language models be too big?}
\newblock In \emph{Proceedings of the 2021 ACM Conference on Fairness, Accountability, and Transparency}, FAccT '21, page 610–623, New York, NY, USA. Association for Computing Machinery.

\bibitem[{Bender and Koller(2020)}]{bender-koller-2020-climbing}
Emily~M. Bender and Alexander Koller. 2020.
\newblock \href {https://doi.org/10.18653/v1/2020.acl-main.463} {Climbing towards {NLU}: {On} meaning, form, and understanding in the age of data}.
\newblock In \emph{Proceedings of the 58th Annual Meeting of the Association for Computational Linguistics}, pages 5185--5198, Online. Association for Computational Linguistics.

\bibitem[{Bybee(2006)}]{Bybee2006}
Joan~L. Bybee. 2006.
\newblock \href {https://doi.org/10.1353/lan.2006.0186} {From usage to grammar: The mind's response to repetition}.
\newblock \emph{Language}, 82(4):711--733.

\bibitem[{Chomsky et~al.(2023)Chomsky, Roberts, and Watumull}]{chomsky_false_2023}
Noam Chomsky, Ian Roberts, and Jeffrey Watumull. 2023.
\newblock \href {https://www.nytimes.com/2023/03/08/opinion/noam-chomsky-chatgpt-ai.html} {The false promise of {ChatGPT}}.
\newblock \emph{The New York Times}.

\bibitem[{Christiansen and Chater(2008)}]{Christiansen2008}
M.~H. Christiansen and N.~Chater. 2008.
\newblock \href {https://doi.org/10.1017/S0140525X08004998} {Language as shaped by the brain}.
\newblock \emph{Behavioral and Brain Sciences}, 31(5):489--558.

\bibitem[{Dębowiak(2014)}]{Debowiak2014}
Przemysław Dębowiak. 2014.
\newblock O dorobku naukowym profesora witolda mańczaka z okazji jubileuszu 90. urodzin.
\newblock \emph{Język Polski}, XCIV(3):194--199.

\bibitem[{Dębowiak(2016)}]{Debowiak_Przemyslaw_Witold_2016}
Przemysław Dębowiak. 2016.
\newblock \href {http://journals.pan.pl/Content/109403/PDF-MASTER/60.00.pdf} {{Witold Mańczak (12 VIII 1924–12 I 2016)}}.
\newblock \emph{Onomastica}, (No 60):5--10.

\bibitem[{Ellis(2002)}]{Ellis_2002}
Nick~C. Ellis. 2002.
\newblock \href {https://doi.org/10.1017/S0272263102002024} {Frequency effects in language processing: A review with implications for theories of implicit and explicit language acquisition}.
\newblock \emph{Studies in Second Language Acquisition}, 24(2):143–188.

\bibitem[{Fló et~al.(2025)Fló, Benjamin, Palu, and Dehaene-Lambertz}]{10.7554/eLife.101802}
Ana Fló, Lucas Benjamin, Marie Palu, and Ghislaine Dehaene-Lambertz. 2025.
\newblock \href {https://doi.org/10.7554/eLife.101802} {Statistical learning beyond words in human neonates}.
\newblock \emph{eLife}, 13:RP101802.

\bibitem[{Fox and Katzir(2024)}]{FoxKatzir}
Danny Fox and Roni Katzir. 2024.
\newblock \href {https://doi.org/doi:10.1515/tl-2024-2005} {Large language models and theoretical linguistics}.
\newblock \emph{Theoretical Linguistics}, 50(1-2):71--76.

\bibitem[{Goldberg(2024)}]{jbp:/content/journals/10.1075/cf.23017.gol}
Adele~E. Goldberg. 2024.
\newblock \href {https://doi.org/10.1075/cf.23017.gol} {Usage-based constructionist approaches and large language models}.
\newblock \emph{Constructions and Frames}, 16(2):220--254.

\bibitem[{Gong et~al.(2018)Gong, He, Tan, Qin, Wang, and Liu}]{DBLP:journals/corr/abs-1809-06858}
ChengYue Gong, Di~He, Xu~Tan, Tao Qin, Liwei Wang, and Tie{-}Yan Liu. 2018.
\newblock \href {https://arxiv.org/abs/1809.06858} {{FRAGE:} {F}requency-agnostic word representation}.
\newblock \emph{CoRR}, abs/1809.06858.

\bibitem[{Grochowski(2017)}]{Grochowski_2017}
Maciej Grochowski. 2017.
\newblock \href {https://doi.org/10.12797/LV.12.2017.2SP.02} {O poglądach profesora {Witolda Mańczaka} na paradygmaty językoznawstwa}.
\newblock \emph{LingVaria}, 12(spec):19--28.

\bibitem[{Hoffmann et~al.(2022)Hoffmann, Borgeaud, Mensch, Buchatskaya, Cai, Rutherford, de~Las~Casas, Hendricks, Welbl, Clark, Hennigan, Noland, Millican, van~den Driessche, Damoc, Guy, Osindero, Simonyan, Elsen, Rae, Vinyals, and Sifre}]{hoffmann2022trainingcomputeoptimallargelanguage}
Jordan Hoffmann, Sebastian Borgeaud, Arthur Mensch, Elena Buchatskaya, Trevor Cai, Eliza Rutherford, Diego de~Las~Casas, Lisa~Anne Hendricks, Johannes Welbl, Aidan Clark, Tom Hennigan, Eric Noland, Katie Millican, George van~den Driessche, Bogdan Damoc, Aurelia Guy, Simon Osindero, Karen Simonyan, Erich Elsen, and 3 others. 2022.
\newblock \href {https://arxiv.org/abs/2203.15556} {Training compute-optimal large language models}.
\newblock \emph{Preprint}, arXiv:2203.15556.

\bibitem[{Ibbotson and Tomasello(2016)}]{Ibbotson2016}
Paul Ibbotson and Michael Tomasello. 2016.
\newblock \href {https://www.scientificamerican.com/article/evidence-rebuts-chomsky-s-theory-of-language-learning/} {Evidence rebuts {Chomsky}'s theory of language learning}.
\newblock \emph{Scientific American}.

\bibitem[{Jaynes(2003)}]{jaynes2003probability}
E.T. Jaynes. 2003.
\newblock \emph{Probability theory: The logic of science}.
\newblock Cambridge University Press.

\bibitem[{Kaplan et~al.(2020)Kaplan, McCandlish, Henighan, Brown, Chess, Child, Gray, Radford, Wu, and Amodei}]{kaplan2020scalinglawsneurallanguage}
Jared Kaplan, Sam McCandlish, Tom Henighan, Tom~B. Brown, Benjamin Chess, Rewon Child, Scott Gray, Alec Radford, Jeffrey Wu, and Dario Amodei. 2020.
\newblock \href {https://arxiv.org/abs/2001.08361} {Scaling laws for neural language models}.
\newblock \emph{Preprint}, arXiv:2001.08361.

\bibitem[{Mandelkern and Linzen(2024)}]{mandelkern-linzen-2024-language}
Matthew Mandelkern and Tal Linzen. 2024.
\newblock \href {https://doi.org/10.1162/coli_a_00522} {Do language models' words refer?}
\newblock \emph{Computational Linguistics}, 50(3):1191--1200.

\bibitem[{Mańczak(1958)}]{Manczak1958Tendances}
Witold Mańczak. 1958.
\newblock Tendances g{\'e}n{\'e}rales des changements analogiques.
\newblock \emph{Lingua}, 7:298--325, 387--420.

\bibitem[{Mańczak(1961)}]{Manczak1961}
Witold Mańczak. 1961.
\newblock O racjonalny dobór haseł w słownikach.
\newblock \emph{Poradnik Językowy}, pages 471--476.

\bibitem[{Mańczak(1969{\natexlab{a}})}]{critique}
Witold Mańczak. 1969{\natexlab{a}}.
\newblock \href {https://doi.org/doi:10.1515/flin.1969.3.3-4.169} {Critique du structuralisme}.
\newblock \emph{Folia Linguistica}, 3(3-4):169--177.

\bibitem[{Mańczak(1969{\natexlab{b}})}]{alma991018253099705524}
Witold Mańczak. 1969{\natexlab{b}}.
\newblock Le développement phonétique des langues romanes et la fréquence.
\newblock Zeszyty naukowe Uniwersytetu Jagiellońskiego 205, Kraków. Nakładem Uniwersytetu Jagiellónskiego.

\bibitem[{Mańczak(1969{\natexlab{c}})}]{Manczak1969a}
Witold Mańczak. 1969{\natexlab{c}}.
\newblock Nieregularny rozwój fonetyczny spowodowany częstością użycia w prasłowiańskim.
\newblock \emph{Slavia}, 38:52--62.

\bibitem[{Mańczak(1978)}]{Manczak1978Analogique}
Witold Mańczak. 1978.
\newblock Les lois du d{\'e}veloppement analogique.
\newblock \emph{Linguistics}, 205:53--60.

\bibitem[{Mańczak(1980)}]{Manczak1980}
Witold Mańczak. 1980.
\newblock Crit{\`e}res de v{\'e}rit{\'e} dans la linguistique.
\newblock \emph{General Linguistics}, 20:140--145.

\bibitem[{Mańczak(1981)}]{Manczak1981KryteriaOrg}
Witold Mańczak. 1981.
\newblock Kryteria prawdy w językoznawstwie.
\newblock \emph{Biuletyn Polskiego Towarzystwa Językoznawczego}, 38:135--142.

\bibitem[{Mańczak(1982)}]{Manczak1982}
Witold Mańczak. 1982.
\newblock Linguistique et autres sciences.
\newblock \emph{Biuletyn Polskiego Towarzystwa J{\k{e}}zykoznawczego}, 39:147--152.

\bibitem[{Mańczak(1987)}]{Manczak1987}
Witold Mańczak. 1987.
\newblock \emph{Frequenzbedingter unregelmäßiger Lautwandel in den germanischen Sprachen}.
\newblock Ossolineum, Wrocław.

\bibitem[{Mańczak(1988{\natexlab{a}})}]{Manczak1988}
Witold Mańczak. 1988{\natexlab{a}}.
\newblock Crit{\`e}res de v{\'e}rit{\'e}. leurs cons{\'e}quences pour la linguistique.
\newblock \emph{Langages}, 89:51--64.

\bibitem[{Mańczak(1988{\natexlab{b}})}]{Manczak1988a}
Witold Mańczak. 1988{\natexlab{b}}.
\newblock O nieregularnym rozwoju fonetycznym spowodowanym frekwencją.
\newblock \emph{Biuletyn Polskiego Towarzystwa Językoznawczego}, 41:105--111.

\bibitem[{Mańczak(1996{\natexlab{a}})}]{Manczak1996Gramatyka}
Witold Mańczak. 1996{\natexlab{a}}.
\newblock Gramatyka opisowa.
\newblock In \emph{Problemy językoznawstwa ogólnego}, pages 139--146. Wrocław.

\bibitem[{Mańczak(1996{\natexlab{b}})}]{Manczak1996GramatykaTransformacyjno}
Witold Mańczak. 1996{\natexlab{b}}.
\newblock Gramatyka transformacyjno-generatywna.
\newblock In \emph{Problemy językoznawstwa ogólnego}, pages 183--190. Wrocław.

\bibitem[{Mańczak(1996{\natexlab{c}})}]{Manczak1996Kryteria}
Witold Mańczak. 1996{\natexlab{c}}.
\newblock Największy problem językoznawstwa: kryteria prawdy.
\newblock In \emph{Problemy językoznawstwa ogólnego}, pages 13--19. Wrocław.

\bibitem[{Mańczak(1996{\natexlab{d}})}]{Manczak1996Nieregularny}
Witold Mańczak. 1996{\natexlab{d}}.
\newblock Nieregularny rozwój fonetyczny spowodowany frekwencją.
\newblock In \emph{Problemy językoznawstwa ogólnego}, pages 52--76. Wrocław.

\bibitem[{Mańczak(1996{\natexlab{e}})}]{Manczak1996Analogicznego}
Witold Mańczak. 1996{\natexlab{e}}.
\newblock Prawa rozwoju analogicznego.
\newblock In \emph{Problemy językoznawstwa ogólnego}, pages 81--97. Wrocław.

\bibitem[{Mańczak(1996{\natexlab{f}})}]{Manczak1996Racjonalny}
Witold Mańczak. 1996{\natexlab{f}}.
\newblock Racjonalny dobór haseł w słownikach.
\newblock In \emph{Problemy językoznawstwa ogólnego}, pages 147--149. Zakład Narodowy im. Ossolińskich - Wydawnictwo, Wrocław.

\bibitem[{Mańczak(1996{\natexlab{g}})}]{Manczak1996Semantyczny}
Witold Mańczak. 1996{\natexlab{g}}.
\newblock Rozwój semantyczny a frekwencja.
\newblock In \emph{Problemy językoznawstwa ogólnego}, pages 121--127. Wrocław.

\bibitem[{Mańczak(1996{\natexlab{h}})}]{Manczak1996Slownik}
Witold Mańczak. 1996{\natexlab{h}}.
\newblock Słownik a gramatyka.
\newblock In \emph{Problemy językoznawstwa ogólnego}, pages 128--132. Zakład Narodowy im. Ossolińskich - Wydawnictwo, Wrocław.

\bibitem[{Piantadosi(2024)}]{piantadosi2023modern}
Steven~T. Piantadosi. 2024.
\newblock \href {https://lingbuzz.net/lingbuzz/007180} {Modern language models refute chomsky's approach to language}.
\newblock In Edward Gibson and Moshe Poliak, editors, \emph{From fieldwork to linguistic theory: {A tribute to Dan Everett (Empirically Oriented Theoretical Morphology and Syntax 15)}}, pages 353---414. Berlin: Language Science Press.

\bibitem[{Piantadosi and Hill(2022)}]{piantadosi2022meaningreferencelargelanguage}
Steven~T. Piantadosi and Felix Hill. 2022.
\newblock \href {https://arxiv.org/abs/2208.02957} {Meaning without reference in large language models}.
\newblock \emph{Preprint}, arXiv:2208.02957.

\bibitem[{Pullum and Scholz(2002)}]{Pullum2002}
G.~K. Pullum and B.~C. Scholz. 2002.
\newblock \href {https://doi.org/10.1515/tlir.19.1-2.9} {Empirical assessment of stimulus poverty arguments}.
\newblock \emph{The Linguistic Review}, 19(1-2):9--50.

\bibitem[{Romberg and Saffran(2010)}]{Romberg2010}
Alexa~R Romberg and Jenny~R Saffran. 2010.
\newblock \href {https://doi.org/10.1002/wcs.78} {Statistical learning and language acquisition.}
\newblock \emph{Wiley Interdisciplinary Reviews: Cognitive Science}, 1(6):906--914.

\bibitem[{Saffran et~al.(1996)Saffran, Newport, and Aslin}]{SAFFRAN1996606}
Jenny~R. Saffran, Elissa~L. Newport, and Richard~N. Aslin. 1996.
\newblock \href {https://doi.org/10.1006/jmla.1996.0032} {Word segmentation: The role of distributional cues}.
\newblock \emph{Journal of Memory and Language}, 35(4):606--621.

\bibitem[{Snow(1972)}]{mothers}
Catherine~E. Snow. 1972.
\newblock \href {http://www.jstor.org/stable/1127555} {Mothers' speech to children learning language}.
\newblock \emph{Child Development}, 43(2):549--565.

\bibitem[{Tomasello(2003)}]{tomasello}
Michael Tomasello. 2003.
\newblock \href {http://www.jstor.org/stable/j.ctv26070v8} {\emph{Constructing a language: A usage-based theory of language Acquisition}}.
\newblock Harvard University Press.

\bibitem[{Zhou et~al.(2021)Zhou, Ethayarajh, and Jurafsky}]{zhou2021frequencybaseddistortionscontextualizedword}
Kaitlyn Zhou, Kawin Ethayarajh, and Dan Jurafsky. 2021.
\newblock \href {https://arxiv.org/abs/2104.08465} {Frequency-based distortions in contextualized word embeddings}.
\newblock \emph{Preprint}, arXiv:2104.08465.

\end{thebibliography}

\end{document}